\documentclass[review]{elsarticle}
\journal{}
\biboptions{authoryear}

\usepackage[margin=2.5cm]{geometry}% by courtesy of Mico
\usepackage{comment}
\usepackage[numbered]{bookmark}
\usepackage{amsmath,amssymb,amsthm}
\usepackage{soul}
\usepackage[ruled,vlined]{algorithm2e}

\usepackage{longtable}
\usepackage{xfrac}
\usepackage{color, colortbl}
\usepackage{rotating}
\usepackage{subfigure}
\usepackage{multirow}
\usepackage{array}
\usepackage{graphicx}
\usepackage{pdflscape}
\usepackage[utf8x]{inputenc}
\usepackage{lmodern,textcomp}

\renewcommand\appendix{\par
  \setcounter{section}{0}
  \setcounter{subsection}{0}
  \setcounter{figure}{0}
  \setcounter{table}{0}
  \renewcommand\thesection{Appendix \Alph{section}}
  \renewcommand\thefigure{\Alph{section}\arabic{figure}}
  \renewcommand\thetable{\Alph{section}\arabic{table}}
}
\theoremstyle{plain}
\theoremstyle{plain}\newtheorem{theo}{{\bf Theorem}}[section]
\theoremstyle{plain}
\usepackage{hyperref}
\hypersetup{
  colorlinks = true, %Colours links instead of ugly boxes
  urlcolor  = blue, %Colour for external hyperlinks
  linkcolor = blue, %Colour of internal links
  citecolor = blue!70, %Colour of citations
  pdfauthor = author
}
\usepackage[title]{appendix}

\begin{document}
\begin{frontmatter}

\title{A novel decision fusion approach for sale price prediction using Elastic Net and MOPSO}
\author[Amiraddress]{Amir Eshaghi Chaleshtori\href{https://orcid.org/0000-0002-5495-5748}{\includegraphics[scale=0.6]{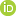}}}
\ead{eshaghi-ch@email.kntu.ac.ir}

\address[AmirAddress]{School of Industrial Engineering, K.N. Toosi University of technology, Tehran, Iran}
\begin{abstract}
Price prediction algorithms propose prices for every product or service according to market trends, projected demand, and other characteristics, including government rules, international transactions, and speculation and expectation. As the dependent variable in price prediction, it is affected by several independent and correlated variables which may challenge the price prediction. In order to overcome this challenge, machine learning  algorithms allow more accurate price prediction without explicitly modeling the relatedness between variables. However, as inputs increase, it challenges the existing machine learning approaches regarding computing efficiency and prediction effectiveness. Hence, this study introduces a novel decision-level fusion approach to select informative variables in price prediction. The suggested meta-heuristic algorithm balances two competitive objective functions, which are defined to improve the prediction’s utilized variables and reduce the error rate simultaneously. In order to generate Pareto-optimal solutions, an Elastic net approach is employed to eliminate unrelated and redundant variables to increase the accuracy. Afterward, we propose a novel method for combining solutions and ensuring that a subset of features is optimal. Two various real data sets evaluate the proposed price prediction method. The results support the suggested model’s superiority concerning its relative root mean square error and adjusted correlation coefficient.
\end{abstract}

\end{frontmatter}

\section{Introduction} \label{section.Introduction}
Intelligent price prediction is a vital function of machine learning and pattern recognition techniques \citep{zhang2018novel}. Researchers have sought to establish more accurate price prediction models to enhance price prediction accuracy. Methodologies range from mathematical models such as linear regression and statistical analysis \citep{ma2016formalized,bhargava2017predicting,chandrasekaran2021forecasting,amik2021application,cho2021genetic,zheng2021share} to alternative techniques employing artificial intelligence, such as expert systems \citep{kim2010preliminary,kim2005hybrid}), neural networks \citep{hsu2011hybrid,kim2005hybrid,bhargava2017predicting,juszczyk2017challenges,odeck2019variation,hao2020modelling,amik2021application,zhang2021research}, and deep learning \cite{wang2020deep,haq2021forecasting,zheng2021share,critien2022bitcoin}.Hence, forecasting prices is essential for every stakeholder in the business to mitigate risks and make well-informed decisions.\\
Many characteristics are usually considered when a price prediction is made using machine learning techniques\citep{asghar2021used}. However, it is hard to identify the most relevant characteristics of a product/service to its price and ignore the others in real-life applications. The presence of unrelated and insignificant features complicates the learning process and impacts the reliability of results. As shown in table \ref{tab:SalepriceLiteratureTable} and \cite{li2021day,yun2021prediction}, the prediction accuracy rely heavily on feature selection methods because it provides a powerful solution against the dimensionally cursed phenomena. By excluding the overlapping and unnecessary features from the data, the important and relevant ones are selected for any data mining process \citep{ayesha2020overview, eesa2013cuttlefish}.\\
% Feature selection methods
Methods for selecting features are initially classified into several categories. The filter, wrapper, and embedded methods are the most prevalent\citep{abd2018role, verleysen2005curse, padmaja2016comparative}, and hybrid \citep{eesa2015novel, leung2008multiple}, and ensemble \citep{lazar2012survey, mahmood2017comparative} are the most recent. In filter methods, candidate subsets are evaluated by using statistics measurements such as mutual information and rough set theory \citep{chandrashekar2014survey, swiniarski2003rough}. Through the use of this approach, features can be selected without relying on any learning models. An added advantage of this technique is its effectiveness and working well with high-dimensional data sets, and it outperforms wrapper methods.
Moreover, using this method has the principal disadvantage of neglecting the selected subsets' integration and the learning models' performance\citep{abd2018role, jindal2017review, teng2003combining}. The wrapper methods assess the quality of features after the most valuable ones are chosen  for prediction. It is more accurate than the filter methods \citep{jain2018feature,jindal2017review,zhao2013cost}. The principal disadvantages of the wrapper methods include computing complexity and greater exposure to over-fitting. In general, wrapper methods require large amounts of computation time to achieve convergence and can be intractable for more extensive data sets \citep{zeebaree2019machine,jain2018feature}. The embedded methods employs a mechanism that guides feature evaluation based on the properties embedded within its learning. Regularization, sometimes referred to as penalization, is the most prevalent embedded method. The embedded methods  eliminate the features with coefficients less than a threshold, and thus the model will be less complex\citep{kabir2010new}. Hence, the embedded method is computationally more efficient and tractable without sacrificing performance than the wrapper and filter methods\citep{cilia2019variable}. Moreover, it can be argued that embedding methods rely on linear correlations between features and response variables, which may not be valid for high-dimensional data sets\cite{amini2021two}.Hence,hybrid methods illustrate recent developments in the feature selection field by combining at least two methods from different approaches (e.g., wrapper, filter, and embedded), procedures that meet the same criteria, or two feature selection methods. A hybrid methodology combines the benefits of both methodologies by combining their complementary characteristics \citep{kabir2010new,peng2010novel}.Moreover, ensemble approaches have recently been developed  \citep{hashemi2021pareto}. These methods derive a subset of features by aggregating the results of several feature selection algorithms. Incorporating numerous procedures allows the methodology for selecting the features to operate better than using one. Moreover, it is possible to treat the feature selection problem as a multi-objective optimization problem \cite{al2020approaches}.However, various multi-objective algorithms may find different feature subsets\citep{vickers2017animal}.Hence, this paper introduces a hybrid procedure to concern this challenge. A multi-objective particle swarm optimization algorithm is utilized in the wrapper approach to lesson the root-mean-square error(RMSE) during the Elastic net(EN) training. Since there is no guarantee that meta-heuristic algorithms achieve the best solution, the EN is implemented to the smaller subset of data to exclude features that are still irrelevant. Then, a new fusion algorithm is developed  to achieve the optimal and stable feature subset. Utilization of the fusion algorithm may lead to achieving a more stable subset of features that guarantee the optimal selected features. The model's efficiency is investigated for predicting sale prices in two real data sets. Following is an outline of the main contributions of this paper:
\begin{itemize}
\item We develop  the fitness functions  to incorporate the number of features and relative RMSE minimization concurrently.
\item We develop a novel decision-level fusion algorithm to achieve a more reliable and stable subset of features.
\item Real case studies of price predictions validate the performance of the proposed method.
\end{itemize}
% Manuscript organisation
The rest of this article is organized in the following manner. Section \ref{section:RelatedWorks} discusses the relevant literature ,and the critical ideas employed in the suggested approach are presented in section \ref{section.MethodAndMaterials}. Section \ref{section.NumericalAnalysis} introduces the sales price data sets and examines the effectiveness of the proposed approach in  price prediction. Finally, section \ref{section.conclusion} concludes the study and makes suggestions for further research.\\
\section{Related Works} \label{section:RelatedWorks}
\subsection{Sale price prediction}\label{Lit.Saleprice}
Studies related to  price prediction have been conducted in several economic areas, such as stock markets\citep{zhang2017stock}, oil prices \citep{cen2019crude}, electricity loads\citep{bicego2018distinctiveness}, and electricity prices \citep{qiu2018ensemble}, real estates \citep{crosby2016spatio}, and airfare \citep{tziridis2017airfare}, which provide essential information for decision making. Hence, the importance of this issue cannot be overstated in the design of today's economic systems. As a result, developing a forecasting prices method is essential for every business stakeholder as it helps reduce risks and make informed business decisions. It is generally possible to classify price prediction methods into two categories based on the task they are intended to perform\citep{ma2020cost}. The first method involves predicting the price trend over a specific time, such as stock and oil prices. Another method involves estimating a particular item's price based on its characteristics, for example, the price of an old automobile or a house. Specifically, this paper discusses the second kind of price prediction task in house and old car price prediction as two real case studies. Table \ref{tab:SalepriceLiteratureTable} reviews the recent studies on price prediction models in various fields. Herein, we investigate all types of price prediction, including cost price, wholesale price, retail price, and sale price, estimated by a data prediction model based on several features. Table \ref{tab:SalepriceLiteratureTable} shows that most studies employ machine learning techniques by considering a few features.\\
\begin{footnotesize}
\begin{landscape}
\begin{longtable}{p{2.5cm}p{9cm}p{1.5cm}p{3cm}p{5cm}}
\caption{An overview of the recent research on the sale price prediction.}\label{tab:SalepriceLiteratureTable}\\
\hline
Author(s)-Year & Research objectives &  No. of Features  & Case study &  Prediction Technique\\
\hline
\cite{kim2005hybrid}&Estimating the initial costs of residential buildings using a hybrid model& 8 & Residential buildings & Neural network and genetic algorithm \\
\cite{kim2010preliminary} &Developing an estimation model for construction costs & 10 & Residential building & Case-based reasoning and genetic algorithm\\
\cite{hsu2011hybrid} & Presenting a hybrid model that can be used to predict stock prices & 16 & Stock price prediction & Self-organizing feature map and genetic programming \\
\cite{gao2013markov} & Designing a multi-objective Markov model aimed at minimizing maintenance costs and users' costs & 5 & Maintenance & Markov model \\
\cite{fereshtehnejad2016multiple} & Introducing a methodology for the evaluation of the life-cycle costs of infrastructure subject to multiple hazards & 5 & Infrastructure project & Dynamic programming procedure \\
\cite{ma2016formalized} & Utilizing an ontology to establish a practical method of formalizing construction costs & 6 & Construction buildings & Specification ontology approach\\
\cite{bhargava2017predicting} & Assessing the likelihood of cost escalation and the severity of deviations in infrastructure projects & 34 & Infrastructure projects & Monte Carlo simulation and statistical models\\
\cite{juszczyk2017challenges} & Using artificial neural networks to estimate construction costs in a non-parametric manner & Not defined & Construction projects & Neural network\\
\cite{wang2020deep} & Proposing a multi-hybrid neural network model based on a complex deep learning algorithm for predicting crude oil prices & 12 & Crude oil price & Deep neural network and wavelet transform \\
\cite{hao2020modelling} &Proposing a hybrid model for carbon price forecasting using feature selection and multi-objective optimization & 30 & Carbon price prediction & Extreme machine learning and grasshopper optimization algorithm\\
\cite{haq2021forecasting} & Developing a two-stage stock price prediction model & 44 & Stock price prediction & Deep generative model\\
\cite{amik2021application} & Providing a system to estimate the price of a pre-owned vehicle for a prospective buyer & 10 & Car price prediction & Linear regression,LASSO, decision tree,random forest\\
\cite{cho2021genetic} & Developing a hybrid cryptocurrency price prediction market & 12 & Predicting a stock market index and cryptocurrency price & Genetic algorithms, regression functions\\
\cite{li2021day} & Analyzing the influence of market coupling on the prediction of electricity prices & 62 & Energy prediction & LSTM deep neural network\\
\cite{yun2021prediction} & Developing an enhanced feature engineering process for stock price direction prediction is proposed & 67 & Stock price prediction & GA-XGBoost algorithm\\
\cite{zhang2021research} & Developing a learning model for accurate evaluation of used car prices& 19& Car price prediction & Support vector regression\\
\cite{zheng2021share} & Using advanced machine learning techniques to improve the accuracy of share price predictions in the aerospace industry & 15 & Predictions for aerospace share prices & Recurrent Neural Networks and principle component analysis \\
\cite{critien2022bitcoin} & Proposing a sentiment analysis approach to predict the Bitcoin price & 5 & Cryptocurrency price prediction & Convolutional neural networks\\
\hline
\end{longtable}
\end{landscape}
\end{footnotesize}
\subsection{Home price prediction}\label{subsec:Lit-Home}
Human settlement is characterized by the price of houses being highly correlated with socioeconomic activities \citep{chen2016challenges}. Researchers and stakeholders in residential construction, urban design, and infrastructure planning can benefit from being aware of the changes in housing prices in addition to house buyers\citep{kang2021understanding}. Researchers have made considerable efforts in economics, econometrics, geography, and public policy to understand how property values affect socioeconomic environments over the last several decades\citep{archer1996measuring,cao2019big,fu2019street,hu2019monitoring}. Hence, house prices increase yearly due to increased  demand\citep{zulkifley2020house}. The house-price model provides numerous benefits for home buyers, investors, and builders. Through this model, home-buyers, lenders, and builders will access valuable information, such as current home prices, enabling them to estimate a house's price. Additionally, this model can assist purchasers in choosing a house based on their budget\citep{febrita2017data}. As a result, it is imperative to develop a new predictive model that estimates the house price with the highest accuracy and reliability. A problem arises when the house price is affected by multiple factors, such as location and property demand. Therefore, most stakeholders, including purchasers, investors, contractors, and the real estate industry, are interested in knowing the exact factors affecting house prices, enabling investors to make informed decisions and assist builders in determining the house price. \\
A paradigm for evaluating housing prices can be classified as conventional or advanced, according to \citep{pagourtzi2003real}. Some traditional methods include step-wise analysis and multiple regression, while recent methods include hedonic pricing tools, artificial neural networks, and spatial analysis \citep{zulkifley2020house}.\\
Various models are available for predicting house prices, so it is essential to select the best model. Among the studies in this field, regression analysis is widely used in several studies \citep{alfiyatin2017modeling,chang2019analysis}
. Support Vector Regression is another standard model for predicting house prices \citep{lin2011predicting,chen2017forecasting,phan2018housing}. Since house price prediction involves many non-linear features, artificial neural networks can model complex non-linear relationships. Hence, \citep{mukhlishin2017predicting,wu2018influence} provide an analysis of home price prediction by artificial neural networks that yields a positive result. In this regard, it can provide an exact prediction model.
Moreover, advanced computer vision techniques have enabled extracting information from images of urban environments. Recently, several studies have demonstrated the effectiveness of visual data in predicting house prices and analyzing neighborhood culture and socioeconomic characteristics \citep{gebru2017using,you2017image,yao2018mapping,law2019take,fu2019street,chen2020measuring,zhang2020uncovering}. \\
\subsection{Old car price prediction}\label{subsec:Lit-Car}
Vehicles in  different categories, like cars, sedans, coupes, support cars, station wagons, hatchbacks, convertible sport-utility vehicles (SUV), minivans, and Pick-up trucks, play a crucial role in human lives\citep{berry2004differentiated}. Moreover, the growing demand for private cars has increased the demand for used cars, creating a market opportunity for both the buyer and seller. Many countries recommend that customers purchase a used car as the price is reasonable and affordable\citep{monburinon2018prediction}. As they are used for a few years, reselling them at a profit may be possible. Nevertheless, several factors contribute to the price of a used vehicle, including its age and condition. Generally, used car pricing does not remain constant. As a result, a predictive model for forecasting car prices is necessary as a supporting tool in the trading process\citep{gajera2021old}.
Predicting used car prices has been the subject of several previous studies. The literature presents a variety of price-prediction models, including multiple linear regression\citep{noor2017vehicle}, support vector regression\citep{peerun2015predicting}, and gradient-boosted regression trees\citep{sun2017price}. Moreover, each prediction model is based on data collected from an online marketplace to identify the most accurate model for predicting used car prices\citep{struyf19cochrane}.\\
\subsection{An overview of feature selection methods}\label{subsec:FeatureSelectionReview}
The pricing of old cars or houses using machine learning directly relates to the process by which technical systems acquire their information, resulting a high-dimensional data \citep{camero2018evolutionary}. Thus, when the information has a significant number of features, the precision of the prediction model and its execution time might be adversely affected, known as the dimensionality curse \citep{gegic2019car}. However, a variety of these features are partial/inappropriate or irrelevant to the sale price \citep{myers2016construction,rafiei2018novel}. Therefore, if the information consists of excessive features, resulting (1) the requirement of ample storage capacity and (2) decreased prediction speed and accuracy. Thus, preprocessing of data aimed at dimensionality reduction is crucial. In this vein, instance and feature selection models are widely used \citep{perez2015simultaneous}.\\
Feature selection is an NP-hard problem with $2^{n}$ states, where $n$ is the number of features. It is becoming increasingly complex as $n$ increases\cite{al2020approaches}.
Our investigation of feature selection literature demonstrates that most studies focus on classification problems (see Table \ref{tab:LiteratureTable}, last columns). That is maybe  most real-world classification problems often lack sufficient information about relevant features. Thus, numerous features are presented to demonstrate the domain, resulting in redundant or irrelevant information. It is possible to significantly reduce the number of irrelevant/redundant features while yielding a more general classifier. This is beneficial for understanding the underlying concept of real-world classification issues \citep{tang2014feature}. For classification problems, the choice of features can take place independently of the learning algorithm, such as filter models  \citep{tian2020robust,hoque2018efs,nguyen2016new,xue2013multi} or the evaluation of the quality of selected features by utilizing the learning algorithm, such as wrapper models \citep{kozodoi2019multi,gonzalez2019new}. In embedded models, features are selected during training so that the model becomes less complex by discarding features whose coefficients are below a certain level \citep{zhang2017high,marafino2015efficient,  zheng2011experimental,ma2007supervised}. Hybrid models integrate two or more well-studied models to create a novel approach to solving feature selection challenges in classification \citep{zhou2020many,nguyen2016new, khan2016multi,xue2012particle,ma2007supervised}.Compared with traditional techniques, the hybrid approach frequently capitalizes on the sub-models advantages and is more robust \citep{got2021hybrid,lu2017hybrid,khan2016multi}. In the ensemble models, the output of many feature selection strategies is assembled to get the ultimate feature set \citep{tsai2020ensemble,ng2020training,drotar2019ensemble,ansari2019ensemble,brahim2018ensemble,seijo2017ensemble,das2017ensemble,ebrahimpour2017ensemble}. Moreover, in the hybrid and ensemble models proposed for classification problems, various meta-heuristic algorithms are applied for feature selection problems. NSGA-II \citep{kozodoi2019multi,li2016bi,gonzalez2019new}, MOPSO \citep{amoozegar2018optimizing,ng2020training,xue2012particle}, SPEA-II \citep{xue2013multi}, MOEAD \citep{zhou2020many} are the most common algorithms which applied to solve this issue since they have lower computational cost and convergence more rapidly than other algorithms.
As shown in Table \ref{tab:LiteratureTable}, most  of the hybrid or ensemble-based feature selection models target this problem for classification and  have not been adequately considered in regression problems.This is a major motivation for this study.
It is often possible to describe a regression model using a linear combination of the most informative characteristics using a regression model based on an ordinary least squares (OLS) approach and a statistical regularizer\citep{zhang2017high}. Therefore, regression models retain information regarding the initial feature space and allow us to determine the most relevant ones. The most known embedded models are Lasso \citep{muthukrishnan2016lasso,zhang2017high,zheng2011experimental} and Elastic net(EN) \citep{amini2021two,marafino2015efficient,panagakis2014elastic}, owing to their ability to modify penalty terms within the regularization method and time-savings.
Table \ref{tab:LiteratureTable} also illustrates that the most recent relevant study in this field is by \cite{amini2021two},  in which a two-stage approach with a genetic algorithm is developed in the first stage to eliminate the irrelevant features. Then, the EN algorithm is implemented to omit the redundant features. In this study, the regularization parameters of EN algorithm is tuned manually after the first stage and the authors claim  that the algorithm achieves sub-optimal solutions for any regression problem.
Moreover, in\cite{amini2021two}, the feature selection problem is formulated as a single-objective optimization problem that does not guarantee the best solution. In contrast, in our paper, the regularization parameters are tuned with the MOPSO algorithm simultaneously to guarantee the optimal solution. Hence, the problem is solved by modeling the feature selection problem as a multi-objective optimization problem.  We adopt the MOPSO to boost the search quality since it can achieve superior results and prevent a thorough search for the Pareto feature subsets. Then, to achieve the stable subset of the optimal feature, a new algorithm has been proposed, which lies in getting the final optimal feature subset. In this stage, the proposed method as a decision-level fusion procedure has been applied to merge the Pareto subset of features to achieve a stable and reliable subset. Hence, the proposed method may improve the accuracy and prediction error in regression problems.\\

\newpage
\begin{footnotesize}
\begin{landscape}
\begin{longtable}{p{2.5cm}p{8cm}cccccp{4cm}cc}
\caption{An overview of recent research on feature selection.}\label{tab:LiteratureTable}\\
\hline
Author(s) & Research objectives & \multicolumn{5}{c}{Feature selection approach}  & Case study &  \multicolumn{2}{c}{Problem Type}  \\ \cline{3-7} \cline{9-10}
 &  & \rotatebox{-90}{Filter} & \rotatebox{-90}{Wrapper} & \rotatebox{-90}{Embedding} & \rotatebox{-90}{Hybrid} & \rotatebox{-90}{Ensemble} &  & \rotatebox{-90}{Classification} & \rotatebox{-90}{Regression}\\
\hline
 \cite{ma2007supervised}&Proposing a new Lasso classifier&-&-&\checkmark&\checkmark&-&Micro-array gene expression data&\checkmark&- \\
 \cite{zheng2011experimental}&Developing a new feature selection method based on Lasso algorithm&-&-&\checkmark&-&-&Cancer data&\checkmark&- \\
 \cite{xue2012particle}&Proposing two feature selection methods using MOPSO&\checkmark&-&-&\checkmark&-& UCI data sets&\checkmark&- \\
\cite{xue2013multi}&Designing a filter-based feature selection method using mutual information and entropy metrics&\checkmark&-&-&-&-& UCI data sets&\checkmark&- \\
\cite{panagakis2014elastic}&Developing a new feature selection using the Elastic net regularization method&-&-&\checkmark&-&-&Audio data&-&\checkmark \\
\cite{marafino2015efficient}&Developing an effective feature selection method for text classification&-&-&\checkmark&\checkmark&-&Text data&\checkmark&- \\
\cite{khan2016multi}&Proposing a hybrid feature selection method&-&-&-&\checkmark&-&UCI data sets&\checkmark&- \\
\cite{nguyen2016new}&Proposing two feature selection methods by inserting swapping and non-dominating strategy to improve local search&\checkmark&-&-&\checkmark&-&UCI data sets&\checkmark&- \\
\cite{muthukrishnan2016lasso}&Exploring the performance of the traditional feature selection method and Lasso&-&-&\checkmark&-&-& Real data in R packages&-&\checkmark \\
 \cite{zhang2017high}&Devising a new regularization based on Lasso regression for feature selection problem&-&-&\checkmark&-&-&The USPS handwriting digit data set/UCI data sets&\checkmark&- \\
 \cite{seijo2017ensemble}&Proposing two different homogeneous and non-homogeneous feature selection methods&-&-&-&-&\checkmark&UCI data sets&\checkmark&- \\
 \cite{das2017ensemble}&Proposing a new ensemble feature selection method&-&-&-&-&\checkmark&UCI data sets&\checkmark&- \\
 \cite{ebrahimpour2017ensemble}&Proposing an ensemble feature selection method based on maximum relevance and minimum redundancy&\checkmark&-&-&\checkmark&\checkmark&Micro array high-dimensional data sets&\checkmark&- \\
 \cite{sohrabi2017multi}&Predicting the dose of Warfarin&-&-&-&\checkmark&-&Warfarin data set&-&\checkmark \\
 \cite{hoque2018efs}&Suggesting a new feature selection algorithm using filter feature ranking methods&\checkmark&-&-&-&\checkmark&Network data set /UCI data sets&\checkmark&- \\
\cite{brahim2018ensemble}&Designing an ensemble feature selection method using various features ranker methods&-&-&-&-&\checkmark&Lymphoma data sets&\checkmark&- \\
\cite{ansari2019ensemble}&Proposing an ensemble feature selection method for sentiment classification&-&-&-&-&\checkmark&Movie review  datasets/Amazon product data set&\checkmark&- \\
\cite{kozodoi2019multi}&Proposing a wrapper method based on NSGA-II&-&\checkmark&-&-&-&Credit scoring data&\checkmark&- \\
\cite{gonzalez2019new}&Proposing a new wrapper method based on NSGA-II&-&\checkmark&-&-&-&Motor imagery data&\checkmark&- \\
\cite{drotar2019ensemble}&Proposing several ensemble techniques for feature selection problem&-&-&-&-&\checkmark& Madelon data sets&\checkmark&- \\
\cite{tian2020robust}&Proposing a heterogeneous ensemble approach&\checkmark&-&-&-&\checkmark&Sensor based human activity data&\checkmark&- \\
\cite{zhou2020many}&Designing a two-level strategy method for feature selection problem using MOPSO and MOEA/D&\checkmark&-&-&-&\checkmark&Sensor based human activity data&\checkmark&- \\
\cite{ng2020training}&Proposing a training error and sensitivity based ensemble feature selection method using NSGA-II&-&-&-&-&\checkmark& UCI data sets&\checkmark&- \\
\cite{tsai2020ensemble}&Proposing an ensemble feature selection using parallel techniques&-&-&-&-&\checkmark& UCI data sets&\checkmark&- \\
\cite{amini2021two}&Proposing a new hybrid GA-EN feature selection approach for regression problems&-&-&-&\checkmark&-& US-NAM parent data sets&-&\checkmark \\
\cite{kenney2021mip}&Developing a mixed integer programming feature selection method for regression problems&-&-&-&\checkmark&-& simulated data&-&\checkmark \\
\cite{got2021hybrid}&Devising a novel hybrid filter-wrapper feature selection strategy using the whale optimization algorithm&\checkmark&\checkmark&-&\checkmark&-& UCI data sets&\checkmark&- \\
\cite{hashemi2021pareto}&Proposing a Pareto-based ensemble feature selection method&-&-&-&-&\checkmark& UCI data sets&\checkmark&- \\
\hline
\textbf{This Study}&\textbf{Proposing a decision-level fusion method for Pareto-based ensemble feature selection }&\textbf{-}&\textbf{\checkmark }&\textbf{\checkmark }&\textbf{\checkmark }&\textbf{\checkmark}&\textbf{Two high-dimensional price prediction data sets}&\textbf{-}&\textbf{\checkmark}\\
\hline
\end{longtable}
\end{landscape}
\end{footnotesize}

\section{Model building} \label{section.MethodAndMaterials}
This section describes the proposed approach for selecting features in regression problems. Initially, a wrapper-embedded method is developed to identify the feature set that provides the best prediction accuracy while it consists of the least possible cardinality. Meta-heuristic search strategies and EN are used simultaneously in this process. Through the use of the subset of features determined by the meta-heuristic algorithm, EN has been used to eliminate remaining unrelated features for enhancing forecasting accuracy. The outcome of this step is the subsets of features chosen by the wrapper-embedded approach. Then, we apply a new fusion method to achieve a more accurate and stable optimal solution set. Hence, the fusion algorithm is utilized as a decision-level fusion mechanism to integrate the Pareto-optimal solutions and improve accuracy and error prediction.
\subsection{Elastic Net} \label{subsection.ElasticNet}
The EN is a model that integrates Ridge regression and Lasso regularization. Like Lasso, EN can construct a simplified version of the regression model by making coefficients with zero values. Studies have shown that the EN model can improve upon Lasso on highly correlated data \citep{ref37}. For an $\alpha$  between 0 and 1, and a non-negative $\lambda$, EN solves the following problem \citep{ref38}:
\begin{equation}
\label{Eq_01}
\min_{\beta_{0},\beta}\left({\frac{1}{2N}\sum_{i=1}^{N}{(y_{i}-\beta_{0}-x_{i}^{T}\beta)^2}+\lambda P_{\alpha}(\beta)}\right)
\end{equation}
where,
\begin{equation}
\label{Eq_02}
P_{\alpha}(\beta)=\sum_{j=1}^{P}{\left(\frac{(1-\alpha)}{2}\beta_{j}^2+\alpha \mid \beta_{j} \mid   \right)}
\end{equation}
In this model, $N$ refers to the cardinality of data, $y_{i}$  is the corresponding dependent variable at $i^{th}$ data, $x_{i}$ refers to the $i^{th}$ variable, $\lambda$ corresponds to a value of the regularization parameter, while $\beta_{0}$ and $\beta$ are scalar, and the $p-$vector, respectively. EN produces the same results as Lasso when $\alpha=1$, and once  $\alpha$ shrinks toward $0$; EN becomes closer to Ridge regression; otherwise, the penalty term $P_{\alpha}(\beta) $ ranges between the  $L^{1}$ norm of $\beta$ and the $L^{2}$ norm of $\beta$.\\
\subsection{The MOPSO algorithm} \label{subsection.MOPSO}
Meta-heuristic algorithms can be utilized to select features based on their capabilities. MOPSO is chosen in this article because it is compatible with many problems associated with feature selection. If the objectives are complicated, the MOPSO is an appropriate tool. Following is a description of the suggested algorithm:
\begin{enumerate}
    \item \textit{Encoding and decoding particle:}
     In MOPSO, developing an appropriate coding/decoding approach is crucial to represent a possible solution. Typically, a binary strategy is utilized to create a solution \citep{chuang2011chaotic,unler2010discrete,wang2007feature}.The structure of the proposed solution is regarded as a vector with  $p+1$ elements, where $p$ indicates the number of features,and the last part of the solution is a real number between 0 and 1, denoted by the regularization parameter of EN ($\lambda$). This part of the solution is used for tuning the hyper-parameters of EN. In the decoding stage, the solution $X_{i}$ is converted into a solution$Z_{i}$ as follows:
    \begin{equation}
    Z_{i j}=\left\{\begin{array}{cc}
    1 & x_{i j} \geq 0.5 \\
    0 & \text { otherwise }
    \end{array}\right.
    \end{equation}
    where, $Z_{ij}=1$ reveals that the $j^{th}$ feature is included in the feature subset $Z_{i}$.

    \item \textit{Fitness evaluation:} This paper regards two objective conflict functions to select the feature sets that provide high prediction accuracy while using a minimum cardinality. Thus, the solutions identified through the search process are archived in a manner that maintains the solutions with the most significant features and the lowest prediction error.
    \item \textit{Update the archive:}This paper uses the crowding distance to determine the number of new individuals. Because the crowding distance has no parameters, this article employs it to eliminate or update the full archive. Several multi-objective evolutionary algorithms have used the crowding distance\citep{xue2012particle, deb2002fast, hamdani2007multi,ref44}.Assuming that the archive at iteration t is At and the new swarm is St. First, we select all solutions that are non-dominated in the combined swarm $P_[t]=A_[t]\bigcup S_{t}$. Then, the non-dominated solutions are added to replace their prior contents. Upon reaching the maximum size of the archive, only those with the most significant crowding distance are preserved (the number of solutions($Na$), also called archive size).
    \item \textit{Update the personal and global best position:}$Pbest$ represents the particle's best position. This paper utilizes a domination-based strategy proposed by \cite{reyes2006multi}to update the particles' best positions. The old particle's memory is kept if the decoded solution is dominant; otherwise, the decoded solution is changed with the old one. $Gbest$ is the best solution derived from the particle neighbors. Due to the conflicting nature of multi-objective optimization problems, selecting the best solution can take time and effort. Hence, based on the diversity of non-dominated solutions in the archive, we employ the crowding distance proposed by \cite{ref41} to evaluate the diversity of non-dominated solutions. Moreover, the binary tournament is applied with these crowding distances to identify each particle's $Gbest$. A higher crowding distance denotes a higher probability of selection as the $Gbest$.
    \item \textit{Mutation Operator:}After exploring the solution space, if some particles locate a suitable position, the others will fly toward that position. This location may be a local optimum, so particles will not re-explore the entire solution space. Furthermore, the algorithm will get stocked into local optima due to a loss of diversity of particles \citep{cheng2011promoting}. Hence, a hybrid mutation based on two efficient operators is presented in this article to avoid the above limitations. The first is the reset operator, intended to preserve the diversity of the swarm. This operator allows some particles to reset their velocity periodically. The other is hop mutation, which enhances searching  in general. As a result of this operator, each particle has a uniform jump probability in any dimensional space. Generally, not all components are requested by the hop operator, which can be regarded as a partial particle reset. Thus, unlike the first operator, it permits more exploration in solution space. Using the two operators in combination does not significantly increase processing time compared to the conventional MOPSO.
    \item \textit{Termination condition:} Stop conditions are a set of criteria satisfying the viability of a solution. This study considers the maximum iteration number for stopping the MOPSO algorithm.
\end{enumerate}

In detail, the proposed algorithm is described in algorithm \ref{alg:MOPSO}.$Pb_{i}(t)=\big(Pb_1(t),Pb_2(t),\ldots,Pb_P(t)\big)$ represents $Pbest$ set and $\big(Gb_1(t),Gb_2(t),\ldots,Gb_P(t)\big)$ is the $Gbest$ set. The coefficients $(c_{1})$ and $(c_{2})$ act as constants that control the effect of $Pbest$ and $Gbest$ within the exploration. $(w)$ serves as an initial weight to handle particle exploration. The $(r_{1})$and $(r_{2})$ are random values between zero to one, and $(v(t))$ is the velocity of the swarm within the search space. For a MOPSO algorithm, \cite{chakraborty2011erratum} introduces its criteria for convergence based on objective functions. In the following, we will now restate the theorem.
\begin{algorithm} \label{alg:MOPSO}
    \SetAlFnt{\small}
    \SetAlCapFnt{\small}
    \SetAlCapNameFnt{\small}
    \footnotesize
    \caption{Pseudo code of the proposed algorithm.}
     \textbf{Initialize:} \\
     Set the parameters, including the swarm size $N_{s}$, the archive size $N_{a}$, the maximum iteration $T_{max}$, and create an empty archive.\\
     \textbf{Evaluate the fitness function:} \\
    Compute the fitness values of all the particles.\\
    \textbf{Update the external archive:}\\
    Use the method presented  in \ref{subsection.MOPSO} and step 3;\\
    \textbf{Update the best positions of particles:}\\
    Using the method introduced in \ref{subsection.MOPSO} and step 4,update the $Pbest$ and $Gbest$ positions ;\\
    Update the particles’ positions using the following equations:
    \begin{equation}\nonumber
        \begin{aligned}
            &v_{ij}(t+1)=w * v_{ij}(t)+r_{1} * c_{1} *\big(Pb_{ij}(t)-x_{ij}(t)\big)+r_{2} * c_{2} *\big(Gb_{j}(t)-x_{ij}(t)\big)\\
            &x_{ij}(t+1)=x_{ij}(t)+v_{ij}(t+1)
        \end{aligned}
     \end{equation}\\
     \textbf{Mutation:}\\
     Apply the proposed mutation operator  and modify part of the particle's velocity using the procedure introduced in section \ref{subsection.MOPSO} and step 5;\\
      \textbf{Termination condition:}\\
      Assess the algorithm's convergence with the termination criteria. The archive should be output as final if it does so. Otherwise, return to the" Update the external archive" step.
\end{algorithm}

\begin{theo}\label{Theo: covergenceOptimality}
Consider $S$ as the initial population. Under the restrictions formulated in equation (\ref{eq.Restriction}), MOPSO will ensure that the mean swarm converges to the Pareto optimal set $\Lambda^{*}$. If $\mu_{t}$ is the mean of the population after $t$ iterations, then $\lim _{t \rightarrow \infty} \mu_{t}=\bar{X}^{*}$ where $\bar{X}^{*}$ denotes the Pareto optimal point.
\begin{equation}\label{eq.Restriction}
w<1 ;\quad 0<\frac{c_{1}+c_{r}}{r}<r(1+w)
\end{equation}
\end{theo}
In the presented algorithm, the inertia weight is adjusted  to  $w = 0.4$.The acceleration coefficients are determined by equation (\ref{eq.PSOcoeff}), where $T_{max}$ is the number of iterations performed. Since these coefficients satisfy equation (\ref{eq.Restriction}); thus, the swarm converges to the center of the Pareto optimal set  $\Lambda^{*}$ as the iteration number $(r_{t})$  grows.
\begin{equation}\label{eq.PSOcoeff}
c_{1}=2.5-\frac{r_{t}}{T_{\max }} \quad, \quad c_{r}=0.5+\frac{r_{t}}{T_{\max }}
\end{equation}
%\section{Proposed method}\label{section.ProposedMethod}
There is evidence that members of the Pareto set may disagree with their choice of features. Hence, the present study proposes a novel method that considers the level of support and the importance of features for prediction accuracy in feature selection. The steps of the proposed approach are explained as a new fusion approach.
\subsection{The proposed fusion approach}\label{subsection.fusionAlg}
In the decision-level fusion process, the conflict arising from multiple models must be addressed in the final decision.
In the feature selection problem,  different models may choose various feature subsets. Hence, it is crucial to integrate their decisions to achieve a more reliable subset of features. This article presents a feature selection method and explains thoroughly for regression problems. In the proposed fusion approach, the adjusted coefficient of determination $(R_{adj}^{2})$ indicates the accuracy level of each Pareto member, and the normalized $R_{adj}^{2}$ indicates the degree of support for each of them. In addition, the extra sum of squares $(ESS)$ assesses the level of uncertainty in prediction accuracy. The $ESS$ assigns the initial weight to each selected feature. Then, we apply a simple additive weighing algorithm to calculate the final weight. The final score of the features is utilized to identify the best ones. Therefore, a subset of features is selected using the given scores calculated by the proposed weighting method. The proposed approach is much more reasonable in handling conflicts of Pareto members in selecting the features by MOPSO for regression problems. The suggested technique contains three parts: the calculation of the degree of support of Pareto members using the normalized $R_{adj}^{2}$, the calculation of the uncertainty in prediction accuracy using $ESS$, and the calculation of the final score of features using the simple additive weighting method.
\subsubsection{The adjusted coefficient of determination}\label{section.R2}
The adjusted coefficient of determination is an extension of $R^{2}$, in which the non-essential explanatory variables are penalized \citep{ramsey2012statistical}. In this way, it is possible to judge whether adding an explanatory variable provides a better fit. This article uses this criterion to determine the weight assigned to each Pareto member in the regression model. The summation of assigned weights must equal one. Therefore, the normalized $R_{adj}^{2}$ is considered as the weight of each Pareto member, formulated as follows:
\begin{equation}\label{eq.R2}
R_{a d j}^{2}=1-\left(\frac{n-1}{n-p}\right)\left(1-R^{2}\right)
\end{equation}

\begin{equation}\label{eq.normalR2}
w_{i}=\frac{(R_{adj}^{2})_{i}}{\sum_{k=1}^{N p} (R_{adj}^{2})_{k}} \quad i=1,2, \ldots, N_{p}
\end{equation}
where $N_{p}$ is the number of Pareto fronts,$n$ is the cardinality of observations, and $p$ is the number of features.

\subsubsection{The extra sum of squares(ESS)}\label{section.ESS}
The  $(ESS)$ is the difference between two regression models’ error sum of squares$(SSE)$. The  $ESS$ measures the marginal reduction in $(SSE)$ when another set of features is included in the model. Hence, this assesses the importance level of features in explaining the response variable. Assume that the regression model contains one feature, $X_{1}$. If we add another feature, $X_{2}$, to the regression model,$(ESS)$ defines the proportional variation explained by $X_{2}$. This can be expressed as $SSR(X_{2} | X_{1})$.Therefore, the extra sum of squares explains the part of $(SSE)$ not explained by the original variable $X_{1}$ \citep{ramsey2012statistical}. The $ESS$ can be considered as the confidence/importance degree for the presence of a specified feature in the regression model, calculated as follows:
\begin{equation}\label{eq.ESS}
\operatorname{ESS}\left(x_{2}\right)=\operatorname{SSR}\left(x_{2} \mid x_{1}\right)=\operatorname{SSE}\left(x_{1}\right)-\operatorname{SSE}\left(x_{2}, x_{1}\right)
\end{equation}
Generally, the formula for calculation of $ESS$ of an specified variable $x_{n}$ can be written as follows:
\begin{equation}\label{eq.ESS1}
\operatorname{ESS}\left(x_{n}\right)=\operatorname{SSR}\left(x_{n} \mid x_{1},...,x_{n-1}\right)=\operatorname{SSE}\left(x_{1},...,x_{n-1}\right)-\operatorname{SSE}\left(x_{1},..., x_{n-1},x_{n}\right)
\end{equation}

\subsubsection{The simple additive weighting method(SAW)}\label{section.SAW}
\begin{figure}
    \centering
    \includegraphics[trim={0.5cm 0.5cm 0.5cm 0.5cm},clip,width=0.8\columnwidth]{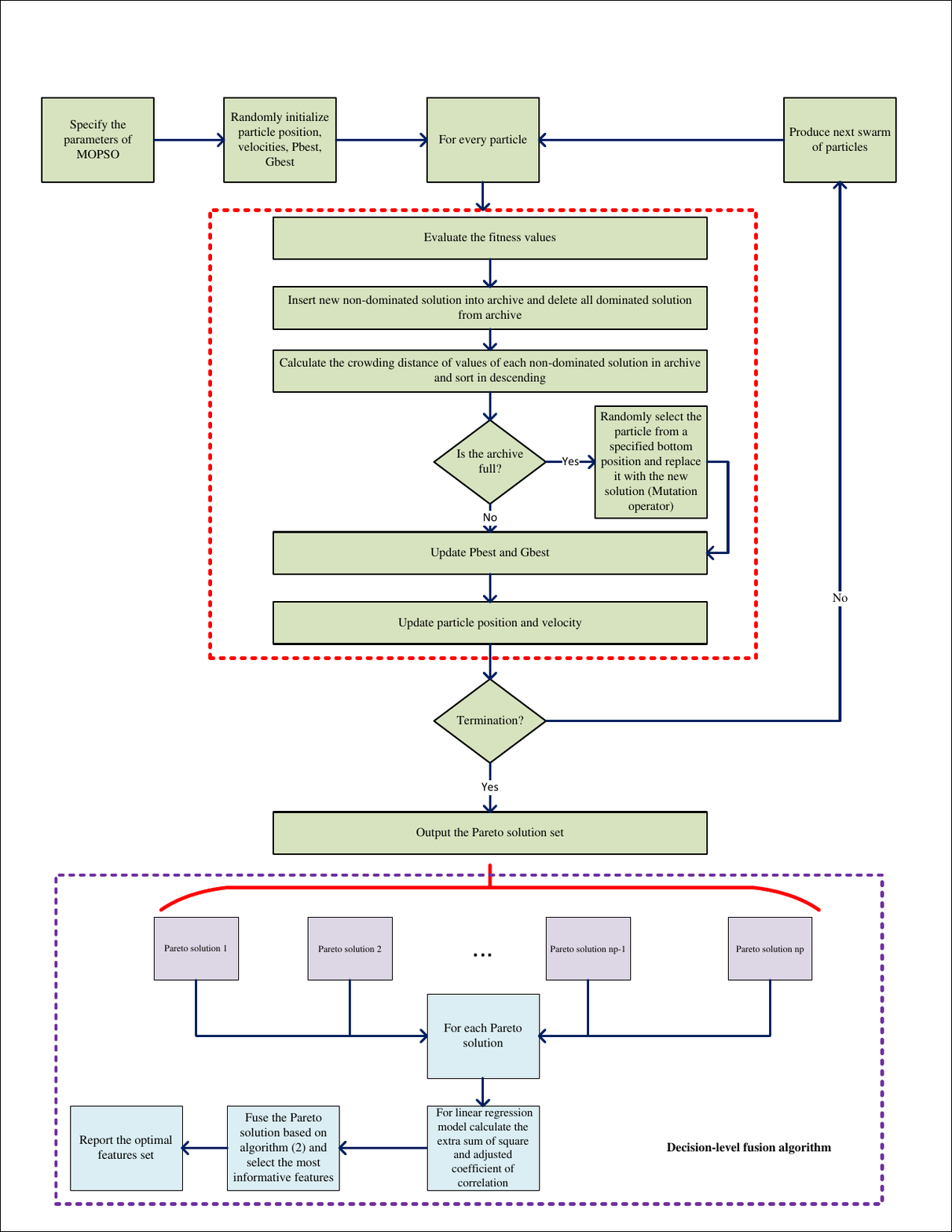}
    \caption{Flow chart of proposed method}
    \label{fig:FlowChart}
\end{figure}

The $(SAW)$ method is the most commonly used $MCDM$ technique, also called the weighted linear combination or scoring method. According to the $SAW$ method, a new weighting approach is suggested to compute the final weight of features. The steps involved in the $SAW$ approach are:
\begin{enumerate}
    \item Define a Feature-Pareto comparison matrix with $(N_{f} *N_{p})$ elements, where $N_{f}$ indicates the cardinality of selected features and $N_{p}$ is the size of Pareto front.
    \item Calculate the weight of each Pareto set according to equation  (\ref{eq.normalR2}).
    \item Compute the $ESS$ related to each feature of the Pareto members, and then assign a score based on equation (\ref{eq.ESS1}) to show how crucial it is. Note that the score of features that do not exist in the specified Pareto member equals zero.
    \item Evaluate the final score of each feature by equation (\ref{eq.FinalScore}):
        \begin{equation}\label{eq.FinalScore}
        \text { Score }_{j}=\sum_{i=1}^{N p} w_{i} ESS_{i j} \quad ; \quad j=1,2, \ldots, N_{f}
        \end{equation}
    \item  High-scoring features are considered members of the final feature set.
\end{enumerate}
Steps involved in the proposed procedure are summarized in algorithm \ref{alg:FusionFe}. The structure of the proposed approach is illustrated in figure(\ref{fig:FlowChart}).
\begin{algorithm} \label{alg:FusionFe}
    \SetAlFnt{\small}
    \SetAlCapFnt{\small}
    \SetAlCapNameFnt{\small}
    \footnotesize
    \caption{Pseudo code of the proposed decision-level fusion approach.}
     \textbf{First stage:} \\
     Run the MOPSO-EN algorithm and find the Pareto sets.\\
     \textbf{Second stage:} \\
     \textbf{Evaluate the importance of features/Pareto sets} \\
    \begin{enumerate}
        \item According to Equation (\ref{eq.normalR2}), calculate the weight of each Pareto member,
        \item Compute the importance of each feature regarding to Equation (\ref{eq.ESS1})
    \end{enumerate}
    \textbf{Decision-level fusion stage:}\\
    According to Section \ref{section.SAW}, implement the SAW method to fuse the Pareto members and obtain the optimal features set
\end{algorithm}

\section{Model evaluation}\label{section.NumericalAnalysis}
This section assesses the suggested feature selection approach. We employ two real price prediction data sets, and tuning the hyper-parameters  should be done before model evaluation.
\subsection{Data sets description }\label{subsection.Dataset}
Predicting the price of houses is crucial for every national economic forecast. We apply the proposed features selection approach to a data set collected for 372 residential apartments built between 1993 and 2008 in Tehran, Iran \cite{ref57,ref58}. The data set contains building costs, sale prices, and project and economic variables to analyze individual residential apartments. The total features equal 105,  and two output variables: construction costs and sale prices. Overall, the number of instances is equal to 372. The presented method has been utilized to evaluate the performance of predicting housing sale prices as the response variable. Because the features have different ranges, they are normalized to the scale range zero to one.\\
%% sale price of used cars
The second data set predicts the selling price of used cars based on their features and current condition. The problem is to model the sale price of used cars using the features provided in the data set. The client could use it to estimate the price of a particular car. This data set includes categorical variables such as the number of seats, fuel type, vehicle type, and owner type.
Moreover, some continuous variables include the year of the model car, mileage driven, the company's standard mileage, the engine displacement volume, and maximum power. Records containing missing values are eliminated first to pre-process the data set. Then, a binary variable is created from the categorical variables, and the continuous variables are converted from zero to one range due to their different ranges. After doing the pre-processing steps, the total features equals 37, and the total instances equal 823.
\subsection{The performance metrics}\label{subsection.PerformaceMetrics}
Since this study's response variables are continuous, the $RMSE_{cv}$ and the cardinality of selected features constitute the evaluation metrics. Moreover, a 10-fold cross-validation process is employed to calculate the first objective. The second one is calculated by counting the cardinality of selected features.

\subsection{Hyper-parameters tuning}\label{subsection.HyperParameter}
It is important to note that the MOPSO has no standard fixed parameters. They significantly impact the MOPSO performance, so they must be adjusted for each problem to achieve the highest exploitation. Theorem \ref{Theo: covergenceOptimality} specifies that the MOPSO must determine the optimal solution early in the search process. As a result, to increase the likelihood of a rapid enhancement in the MOPSO’s response, a high level of elitism, a small swarm size, an archive size, and a pretty high mutation probability are used. The parameters for the MOPSO tuned for this study can be found in table \ref{tab:TableMOPSO}.
\begin{table}[ht]
    \caption{The tuned MOPSO parameters.}
    \label{tab:TableMOPSO}
    \centering
    \begin{tabular}{l c c}
        \hline
        Parameters & \multicolumn{2}{c}{Values/Method} \\
        \hline
         & Car data set& House data set \\
        \hline
         Maximum number of iteration $(T_{max})$& 50&80 \\
         Swarm size $(N_{s})$ & 20&35 \\
         Archive size $(N_{a})$ & 15&20 \\
         Inertia weight $(w)$ & 0.4 &0.4\\
         Initial personal learning coefficient$^{*}$ $(c_{1})$&1&1\\
         Initial global learning coefficient$^{*}$ $(c_{2})$ &2&2\\
         Mutation rate & 0.1&0.1 \\
         Mutation function & \multicolumn{2}{c}{Initialized-jumping operator}\\
         \hline
    \end{tabular}\\
    \footnotesize{$^*$ These hyper-parameters are updated by Equation (\ref{eq.PSOcoeff}) at each iteration}
\end{table}
\subsection{Model validation}\label{subsection.modelvalidation}
This section compares some classical and advanced feature selection algorithms with the proposed method.
\subsubsection{Comparison between the decision-level fusion method and  feature selection method based on MOPSO-EN}
\begin{figure}[ht!]
\begin{center}
\subfigure[House data set ]{
\includegraphics[trim={0.1cm 0.1cm 0.1cm 0.1cm},clip,width=0.4\columnwidth]{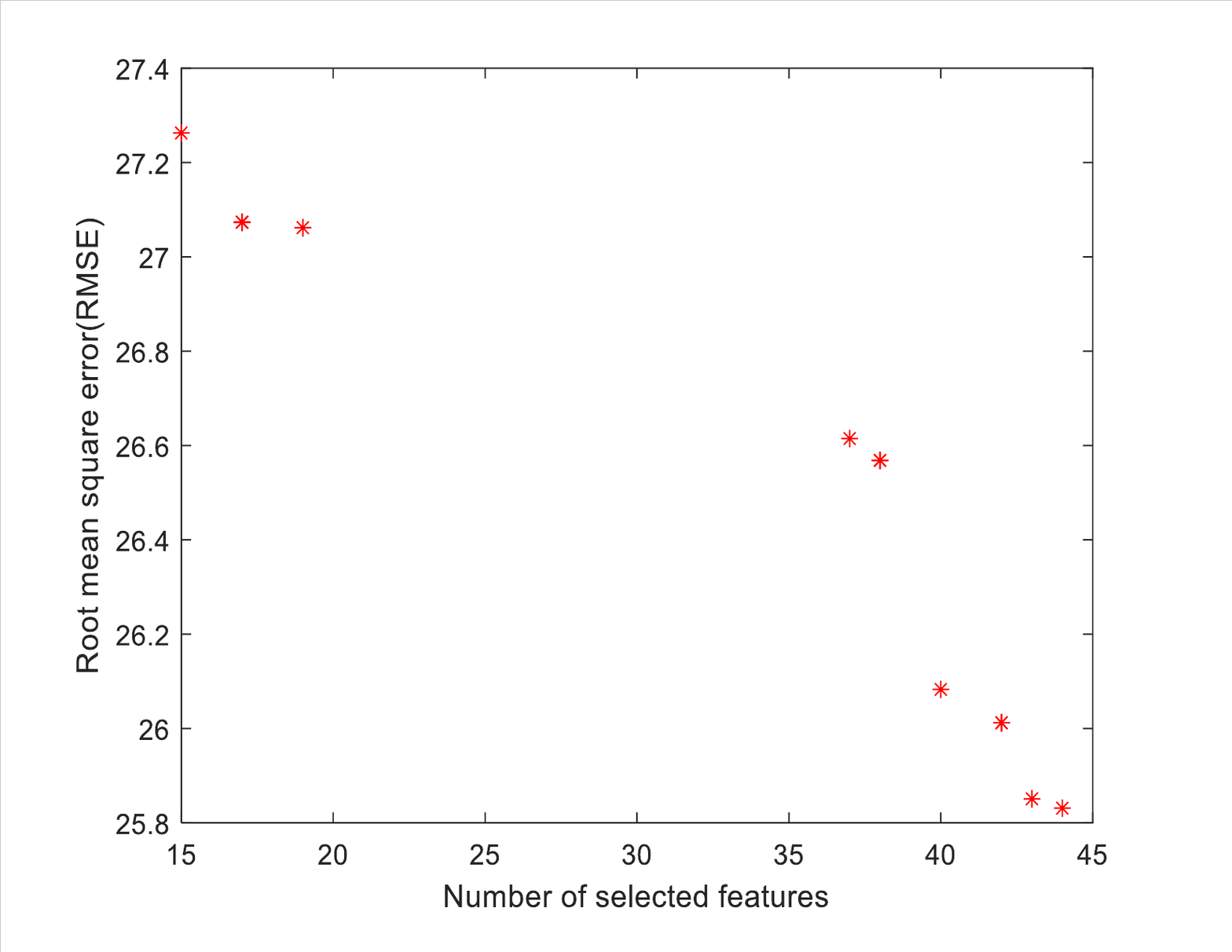}
\label{Fig:HousePareto}
}
\subfigure[Car data set]{
\includegraphics[trim={0.1cm 0.1cm 0.1cm 0.1cm},clip,width=0.4\columnwidth]{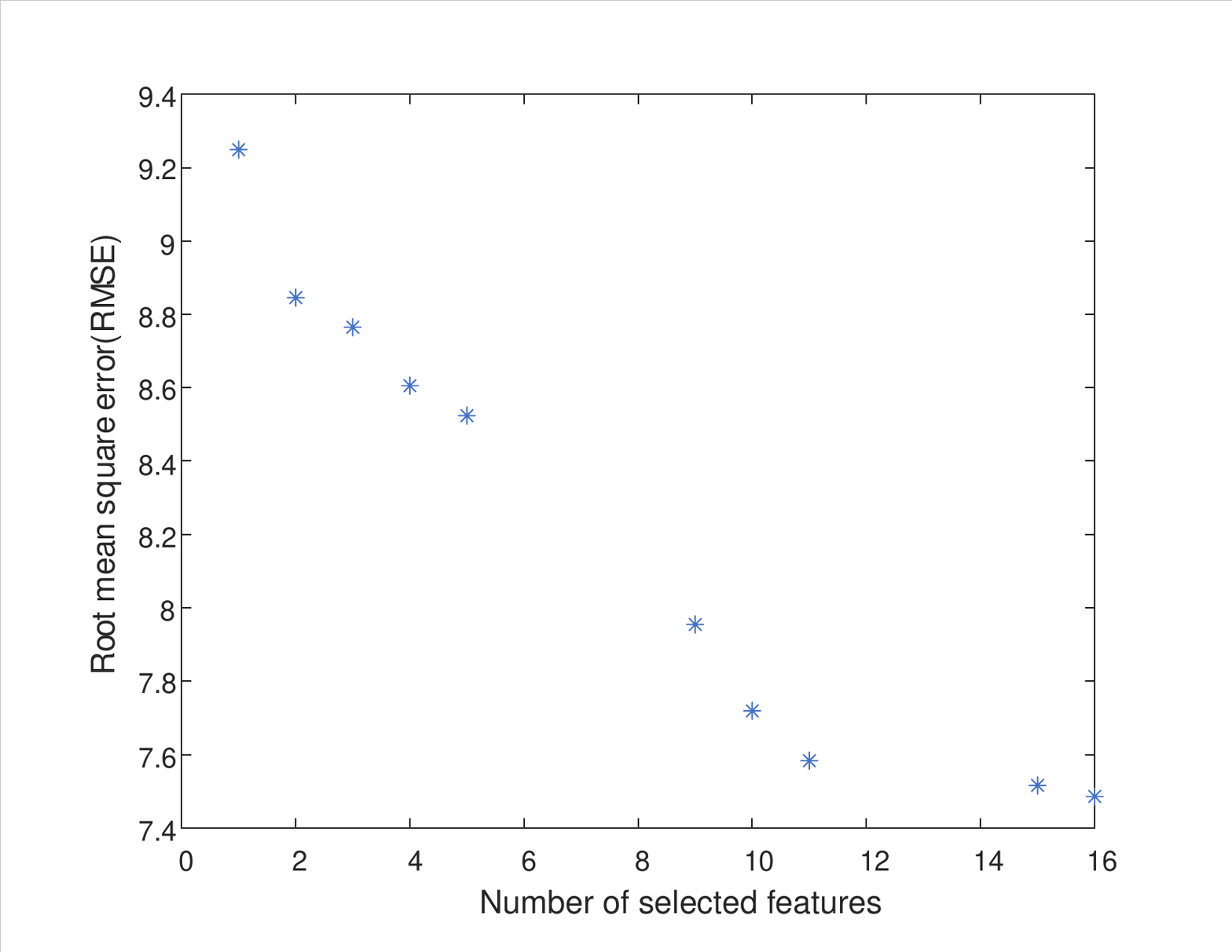}
\label{Fig:CarPareto}
}
\caption{ Pareto set of House data and Car dataset }
\label{Fig:BothParetoSet}
\end{center}
\end{figure}

We evaluate the proposed method by randomly dividing each dataset into two sets, 70\% for training and 30\% for testing. The train data set undergoes a 10-fold cross-validation process to boost its reliability. The proposed method is run in a $MATLAB2020b$ software program using a laptop with a Corei7 processor and 8 GB of RAM. The overall performance of each Pareto and acquired features set merged with algorithm \ref{alg:FusionFe} is evaluated by  $(RMSE_{cv})$ and $R_{adj}^{2}$ metrics. MOPSO-EN is also run for each data set, and the Performance metrics for each Pareto set are reported in table \ref{tab:TableResultsTrain}. The Pareto set for each data set is illustrated in figures\ref{Fig:HousePareto} and \ref{Fig:CarPareto}. According to the performance metrics for the House data set, the sale price of the houses can be predicted accurately by at least 15 independent features (Pareto set 5) up to 40 independent features (Pareto set 8).
Moreover, at each Pareto set, about 98\% of the variability of the response variable is explained by specific characteristics.\\
Regarding the performance metrics for the car data set, the selling price of the used car can be predicted by one independent feature (Pareto set 5) up to 16 independent features (Pareto set 1). Moreover, at each Pareto set, at least 80\% of the variability of the selling price as a response variable can be explained by selected features. As evident in both data sets, the variability of the target variable can be modeled by different feature sets. Hence, to obtain the more reliable independent features set for each data set, algorithm \ref{alg:FusionFe} is applied to the corresponding Pareto features set, and the regression model is built by fused features. According to table \ref{tab:TableResultsFused}, for each benchmark data, the overall performance of the regression model built with the fused features set is higher than the ones in each Pareto set in training data. Furthermore, the model built by fused features can predict the response variable with high accuracy in test data. Our findings indicate that the suggested approach is superior in training and testing data for two benchmark data sets.
\begin{table}[ht]
    \caption{Performance metrics of each Pareto set in MOPSO-EN algorithm for training data.}
    \label{tab:TableResultsTrain}
    \centering
    \begin{tabular}{c c c c| c c c c}
        \hline
        \multicolumn{4}{c}{House dataset} & \multicolumn{4}{c}{Car dataset} \\
        \hline
         Pareto set & \# selected features &$RMSE_{cv}$ &$R_{adj}^{2}(\%)$&Pareto set& \# selected features&$RMSE_{cv}$&$R_{adj}^{2}(\%)$ \\
        \hline
        1&43&25.85&98.48&    1&16&7.48&85.03\\
        2&44&26.13&98.43&    2&15&7.51&84.88\\
        3&17&27.07&98.32&    3&5&8.52&81.49\\
        4&42&26.01&98.47&    4&4&8.60&81.04\\
        5&15&27.10&98.32&    5&1&9.2&79.85\\
        6&38&26.59&98.43&    6&9&7.95&82.92\\
        7&37&26.66&98.41&    7&2&8.84&80.32\\
        8&40&26.08&98.48&    8&3&8.76&80.68\\
        9&19&27.06&98.34&    9&11&7.58&84.73\\
        & & & &              10&10&7.71&84.16\\
    \end{tabular}\\
\end{table}
%-----------------------------------------------------------------------------
\begin{table}[ht]
    \caption{Performance metrics of  the proposed algorithm with fused features set in both benchmark data sets.}
    \label{tab:TableResultsFused}
    \centering
    \begin{tabular}{c| c c c |c c c}
        \hline
        ~  &\multicolumn{3}{c}{House dataset} & \multicolumn{3}{c}{Car dataset} \\
        \hline
         ~&\# fusion features &$RMSE_{cv}$ &$R_{adj}^{2}(\%)$&\# fusion features&$RMSE_{cv}$&$R_{adj}^{2}(\%)$ \\
          \hline
         Train data&\multirow{2}{*}{46}&26.33&98.41&\multirow{2}{*}{15}&6.368&88.45\\
         Test data&~&19.66&98.74&~&7.22&85.38\\
        \hline

    \end{tabular}\\
\end{table}
% figures-------------------------------------------------------------------------------

\subsubsection{Comparison between the decision-level fusion method and  benchmark feature selection methods}
This subsection compares the suggested approach with the Elastic net(EN) and the genetic algorithm combined with linear regression(GA-LR) as benchmarks. In comparison with other meta-heuristic algorithms, GA is a highly adaptable algorithm, which is the primary reason for choosing it as a wrapper method benchmark. Moreover, the GA algorithm employs linear regression without incorporating feature selection. An individual is represented as a binary vector with a length of $P$, where $p_{i}=1$ expresses that the feature pi is selected for the individual, $p_{i}=0$ if the feature $p_{i}$
is absent in the individual $i=(1,2,...,p)$. We generate an initial population consisting of predetermined numbers of individuals and randomly select numbers 0 and 1 for each entry. The solution with the lowest prediction error$(RMSE_{cv})$ and fewer features have been nominated for the next iteration. Equation (\ref{eq.GAFitness}) describes how fitness values for each individual are determined based on the applied fitness function of GA($F_{GA})$, which evaluates the fitness value based on the linear regression model. Also, the weights $w_{r}$ and $w_{p}$   must be adjusted separately for each data set. These coefficients have been applied to assign the weight to each objective function to transform a multi-objective problem into a single objective. In this paper, four scenarios for these weights are considered using a grid search method.\\
\begin{equation}\label{eq.GAFitness}
F_{GA}=w_{r}*RMSE_{cv}+w_{p}*n_{p}
\end{equation}
The elite are chosen as parents according to their fitness value to produce children via crossover and mutation operators. This study's crossover and mutation operators are similar to those presented in \cite{amini2021two}. Because GA parameters are not universally fixed and may impact the GA's performance, it is important to adjust them to the particular problem. Hence, GA parameters can be adjusted to maximize exploitation in this regard. An elitist approach, a small initial population size, and an extremely high mutation probability are applied to ensure a rapid improvement in the GA's performance. Additionally, random individuals are included in each generation to maintain a diverse population.
Moreover, the number of generations must also be kept low to satisfy the time constraint. Table \ref{tab:GaParameters} summarizes the GA parameters tuned in this paper.\\
Several scenarios are presented in table \ref{tab:GAResults}, in which the higher the weight, the greater the importance placed on minimizing the related term. In this study, the scenario with ($(w_{r}=0 \& w_{p}=1)$) is ignored because this study intends to enhance prediction accuracy. Table \ref{tab:GAResults} shows each original and selected number of features related to each scenario. In addition, table \ref{tab:GAResults} shows the performance metrics related to each scenario. For both data sets, the wrapper method shows the most significant feature reduction and the minimum $(RMSE_{cv})$. As a result, this method not only reduces the size and complexity of data but also results in minor predicted errors. However, for the last scenario$(w_{r}=1 \& w_{p}=0)$, GA-LR produces the lowest prediction error, although it specifies the most significant subset. The outcomes in tables \ref{tab:TableResultsFused} and \ref{tab:GAResults} indicate that the suggested featured selection method enhances the prediction model's performance by excluding irrelevant and unnecessary features. Although our proposed method extracts more features than others, resulting in a more accurate prediction and a higher adjusted coefficient for the benchmark data sets. Based on the comparison of performance metrics, the proposed method enhances the prediction accuracy for regression problems.\\
\begin{table}[ht]
    \centering
    \caption{the tuned GA parameters}
    \label{tab:GaParameters}
    \begin{tabular}{c c c}
    \hline
    GA parameters&House data set&Car data set\\
    \cline{2-3}
    ~ &  \multicolumn{2}{c}{Values/Method}\\
    \hline
         number of iterations & 100&80 \\
         Population size & 50&60 \\
         Crossover rate & 0.7&0.8 \\
         Mutation rate & 0.1&0.2 \\
    \hline
    \end{tabular}
\end{table}
%---------------------------------------------------------
\begin{table}[ht]
    \caption{Results of experiments for both data sets in the wrapper method(GA-LR).}
    \label{tab:GAResults}
    \centering
    \begin{tabular}{c|c c c c| c c c}
        \hline
        Scenario&~&\multicolumn{3}{c}{House dataset} & \multicolumn{3}{c}{Car dataset} \\
        \hline
         $(w_{r}-w_{p})$ &&~ \# selected features &$RMSE_{cv}$ &$R_{adj}^{2}(\%)$& \# selected features&$RMSE_{cv}$&$R_{adj}^{2}(\%)$ \\
         \hline
         \multirow{2}{*}{(0.3-0.7)}&train&\multirow{2}{*}{8}&30.51&98.14    &\multirow{2}{*}{1}&10.02&77.43\\
         ~&test&~&29.02&98.35    &~&7.66&79.16\\
        \hline
         \multirow{2}{*}{(0.5-0.5)}&train&\multirow{2}{*}{9}&23.75&98.73    &\multirow{2}{*}{1}&10.04&76.57\\
         ~&test&~&31.57&98.41    &~&7.35&83.06\\

        \hline
       \multirow{2}{*}{(0.7-0.3)}&train&\multirow{2}{*}{12}&27.41&98.62    &\multirow{2}{*}{1}&9.54&78.64\\
         ~&test&~&22.63&98.65    &~&9.04&74.14\\
        \hline
        \multirow{2}{*}{(1-0)}&train&\multirow{2}{*}{62}&19.88&98.99    &\multirow{2}{*}{27}&7.22&84.23\\
         ~&test&~&21.08&98.19    &~&9.67&80.68\\
        \hline

    \end{tabular}\\
\end{table}
In addition, Elastic net has been established as the embedded method since it represents a generalized version of Lasso and Ridge regression. The EN method would achieve optimal results through the careful selection of hyperparameters. Therefore, four regularization parameter values are considered in the grid search subset.
\begin{table}[ht]
    \caption{Results of experiments for both data sets in the embedded method(EN).}
    \label{tab:ENResults}
    \centering
    \begin{tabular}{c|c c c c| c c c}
        \hline
        &~&\multicolumn{3}{c}{House dataset} & \multicolumn{3}{c}{Car dataset} \\
        \hline
         $(\lambda)$ & &~ \# selected features &$RMSE_{cv}$ &$R_{adj}^{2}(\%)$& \# selected features&$RMSE_{cv}$&$R_{adj}^{2}(\%)$ \\
         \hline
         \multirow{2}{*}{0}&train&\multirow{2}{*}{103}&427.2&97.73    &\multirow{2}{*}{33}&203.28&83.85\\
         ~&test&~&278.44&84.72    &~&123.23&77.58\\
        \hline
         \multirow{2}{*}{0.25}&train&\multirow{2}{*}{99}&430.49&97.89    &\multirow{2}{*}{12}&222.24&80\\
         ~&test&~&286.46&80.61   &~&126.6&74.94\\

        \hline
       \multirow{2}{*}{0.5}&train&\multirow{2}{*}{37}&433&97.8    &\multirow{2}{*}{6}&239.27&77.11\\
         ~&test&~&302.11&81.73    &~&134.79&74.99\\
        \hline
        \multirow{2}{*}{1}&train&\multirow{2}{*}{27}&436.28&97.62   &\multirow{2}{*}{4}&268.14&61.28\\
         ~&test&~&310.39&81.11    &~&219.12&52.76\\
        \hline

    \end{tabular}\\
\end{table}
Table \ref{tab:ENResults} shows the $(RMSE_{cv})$ related to each scenario and its selected features. This table shows that increasing the regularization parameter $(\lambda)$ leads to lower selected features and higher $(RMSE_{cv})$ for both benchmark data sets.
Moreover, the highest variability explained by the EN has been achieved when no regularization exists in the embedded model $(\lambda=0)$. Tables \ref{tab:TableResultsFused} and \ref{tab:ENResults} compare the proposed and wrapping methods, showing the proposed method's performance in enhancing the regression's prediction accuracy. As evident, For both data sets, our proposed method can lead to better performance metrics with the performance obtained by the wrapper method.
\subsection{Statistical analysis of results}\label{section.statisticalAnalysis}
In this section, we perform a non-parametric test called Wilcoxon with a confidence level of 0.9 to identify which algorithms are substantially more efficient than the others \cite{jahani2022covid}. The outcomes of the test are provided in table \ref{tab:Wilcoxontest}. The findings demonstrate that the suggested method outperforms the other two benchmark algorithms. Nevertheless, the red values in the table indicate that the presented approach cannot perform as well as the algorithm pair method. However, despite the uncertainty in the benchmark algorithm, all tests have shown that the offered method is superior to the others and appears error-free.\\
\begin{table}[ht]
    \caption{The outputs achieved by the Wilcoxon test for both data sets.}
    \label{tab:Wilcoxontest}
    \centering
    \resizebox{0.95\textwidth}{!}{
    \begin{tabular}{c c|c c |c c}
        \hline
        ~& ~  &\multicolumn{2}{c}{House dataset} & \multicolumn{2}{c}{Car dataset} \\
        \hline
         ~&~&Proposed Alg vs GA-LR &Proposed Alg vs EN&Proposed Alg vs GA-LR&Proposed Alg vs EN \\
          \hline
         \multirow{2}{*}{P-value(Train data-Test data)}&$RMSE_{cv}$&(\textcolor{green}{0.05-0.05})&(\textcolor{green}{0.05-0.05})&(\textcolor{red}{(0.978-0.977})&(\textcolor{green}{0.05-0.05})\\
         ~&$R_{adj}^{2}$&(\textcolor{red}{0.978-0.978})&(\textcolor{red}{0.978}-\textcolor{green}{0.05})&(\textcolor{green}{0.05-0.05})&(\textcolor{green}{0.05-0.05})\\
        \hline

    \end{tabular}}\\
\end{table}
\section{Conclusions and future directions}\label{section.conclusion}
This paper proposes an innovative procedure for selecting the most significant subset of features to enhance regression model accuracy. It includes two stages: a hybrid wrapper embedded method and a decision-level fusion technique. Initially, the MOPSO-EN explores the solution space to identify the subsets of features with the slightest prediction error and minimum cardinality. It can minimize the process of finding the optimal features by circumventing an exhaustive search across all possible solutions. Then, a novel decision-level fusion algorithm is used to obtain more accurate and reliable optimal features, improving accuracy and prediction error and getting more reliable and stable features for a regression problem. The superiority of the presented method has been discussed on two real data sets. According to the outcomes, compared to the other methods, the presented method effectively reduces the number of features and minimum $(RMSE_{cv})$. Hence, combining wrapped features with decision-level fusion reduces the dimension of feature space without compromising accuracy. There are a few limitations to this study that suggest directions for future research. Even though the MOPSO is superior to exhaustive search methods, it still requires much computing power, especially for large data sets.
Moreover, the Elastic net can complement other algorithms to settle this concern in future studies. The suggested method can also be implemented for various data sets not covered in this study. An examination of other fusion-level approaches could be considered for future studies.
\subsection*{\textbf{Acknowledgements}}
This article's research, authorship, or publication does not involve any conflicts of interest on the part of the authors.
\bibliography{Ref}

\end{document}